\documentclass[onecolumn]{musubi}
\usepackage{hyperref}
\usepackage{amssymb}
\usepackage{float}
\usepackage{booktabs}
\usepackage{multirow}

\title{Mela: Test-Time Memory Consolidation based on Transformation Hypothesis}

\author[1]{Lungchuan Chen}

\affiliation[1]{MusubiAI}

\abstract{
Memory consolidation, the process by which transient experiences are transformed into stable, structured representations, is a foundational organizing principle in the human brain, yet it remains largely unexplored as a design principle for modern sequence models. In this work, we leverage established neuroscientific theories of memory consolidation and cross-frequency coupling to propose the Hierarchical Memory Module (HMM), a neural memory architecture composed of two functionally distinct sub-modules that operate at different update frequencies. Inspired by the transformation hypothesis, the low-frequency sub-module produces high-level representations that capture abstract, gist-level knowledge, while the high-frequency sub-module produces fine-grained representations that preserve richer episodic detail. The final memory output is dynamically reconstructed as a context-dependent combination of both representations, analogous to the reconstructive nature of human memory retrieval. We integrate HMM into a Transformer-based language decoder to form Mela, a family of memory-augmented language models that perform online memory consolidation at test time. To further exploit the multi-granularity memory representations produced by HMM, we introduce MemStack, a method that distributes different levels of memory features across the early layers of the decoder without introducing additional tokens. Experiments on language modeling demonstrate that Mela outperforms Transformer baselines across all the model sizes. Moreover, with the pretrained context length fixed at 4K, Mela maintains performance on significantly longer contexts, whereas Transformer baselines degrade rapidly beyond their training length. Extensive ablation studies validate the contribution of each component and provide guidance for practical configuration. Our code is publicly available at \url{https://github.com/Musubi-ai/Mela}.
}

\date{May 2026}
\correspondence{\email{blaze7451@gmail.com}}

\begin{document}
\maketitle

\section{Introduction}
With the recent advances in AI, Transformer \citep{vaswani2017attention} has been adopted across a wide range of domains, including computer vision \citep{dosovitskiy2020image}, natural language processing \citep{radford2018improving}, and time-series modeling \citep{nie2022time}, due to its powerful modeling capability and scalability. The main contributor to its success lies in its core building block, the attention module, which projects each input discrete token into queries, keys, and values and leverages them to compute weighted combinations of values. Through this design, the model is able to assign different levels of importance to input tokens based on their relevance to each other. By aggregating information from all tokens according to the learned attention weights, the model effectively captures long-range inter-token dependencies and constructs context-rich representations that encode contextual information. 

Despite its effectiveness in sequence modeling, the standard implementation of the attention mechanism is well known to incur $O(n^{2})$ time and space complexity, where $n$ denotes the sequence length. As a result, both memory usage and computational cost grow quadratically with increasing sequence length, which significantly hinders the applicability of Transformer models to long-context scenarios. This limitation has motivated extensive research on more efficient attention mechanism and Transformer alternatives \citep{katharopoulos2020transformers, dao2024transformers, yang2024gated, sun2024learning, behrouz2024titans}. Among them, Test-Time Training (TTT) \citep{sun2024learning} has emerged as a promising research direction that divides the whole gradient flow into inner-loop and outer-loop. The inner-loop parameters are updated via online meta-learning, allowing adaptation even at test time, while the outer-loop parameters follow the conventional paradigm of being updated during training and frozen at inference. Titans \citep{behrouz2024titans} builds on this framework by interpreting the meta-learned module as a form of neural long-term memory and the inner loop as a process of memorizing context. This memory-centric interpretation is significant because it bridges machine learning and neurophysiology, opening the door to leveraging established theories of memory to inform model design. 

One such connection has already proven fruitful. In neuroscience, cross-frequency coupling, which refers to the synchronization of neural oscillations across different frequency bands, has been identified as a principal mechanism for integrating information across distributed brain regions \citep{kim2016desynchronization, staresina2024coupled}. Empirical evidence further suggests that cross-frequency coupling in frontal areas is strongly correlated with fluid intelligence \citep{pahor2014theta}, whereas abnormal cross-frequency coupling may contribute to the disruption of certain cognitive functions, as observed in schizophrenia \citep{uhlhaas2013high}. Motivated by this neuroscientific finding, nested learning \citep{behrouz2025nested} proposes to treat the optimization process and the learning architecture as two interconnected nested components that together form a unified model, with each component optimizing its own internal objective at a different level and frequency. This line of work exemplifies a broader design philosophy that we adopt in the present study, namely that neural networks composed of functionally specialized modules with effective mechanisms for inter-module coordination are more likely to give rise higher-level intelligence than approaches that rely solely on scaling model architectures or training data. Recent empirical studies corroborate this viewpoint, demonstrating that modular architectures with structured inter-module communication consistently outperform monolithic models of comparable scale on tasks requiring compositional reasoning and generalization \citep{wang2025hierarchical, behrouz2025nested}.

\subsection*{Memory Formation and Consolidation}
Over the past decades, memory mechanisms have played a crucial role in many deep learning models \citep{hochreiter1997long, weston2014memory, burtsev2020memory, zhang2024memory, beck2024xlstm}. At the same time, memory, despite its varying roles across frameworks, is involved in several influential theories, including Tulving's theory \citep{tulving1985memory}, Global Workspace Theory (GWT) \citep{baars1993cognitive, baars2005global}, the Memory Theory of Consciousness (MToC) \citep{budson2022consciousness, budson2025memory}, Higher-Order Theory (HOT) \citep{ledoux2020seeing}, and Integrated Information Theory (IIT) \citep{tononi2016integrated}. Together, these frameworks suggest that memory is not merely an auxiliary component but a foundational organizing principle, which we argue should be explicitly encoded into the architectural design of models. The purpose of memory is not to perfectly restore past events but rather to furnish the central executive with structured references derived from prior experience, enabling complex cognitive processes such as reasoning, planning, and decision-making. Through a process known in the human brain as memory consolidation, constructing these representations entails progressively building hierarchical abstractions of higher-level conceptual knowledge from low-level input features rather than merely encoding the latter.

Specifically, memory consolidation operates at two levels based on scale and duration, namely synaptic consolidation and system consolidation \citep{dudai2004neurobiology}. Synaptic consolidation refers to the stabilization of local synapse connections that participate in memory encoding, typically completing within hours after learning. System consolidation, by contrast, is the gradual reorganization of memory representations across broadly distributed brain regions, typically extending over weeks to years. In the human brain, labile and transient memories initially encoded by the hippocampus are gradually transformed into more permanent representations stored across cortical regions. Crucially, this transformation does not merely relocate the same content but alters memory both quantitatively and qualitatively., yielding representations that are more abstract, generative, and schematic in nature \citep{diekelmann2010memory}. This process is orchestrated through cross-frequency coupling between the hippocampus and cortical regions, which enables the integration of information across neural systems operating at different temporal scales.

Several theories have been proposed to account for how system consolidation unfolds. The standard consolidation theory (SCT) posits that memories are initially encoded in the hippocampus and gradually transferred to the neocortex, eventually becoming independent of the hippocampus \citep{winocur2011memory}. Multiple trace theory (MTT) challenges this view, arguing that episodic memories always retain hippocampal involvement, with each reactivation generating new traces that render the memory more robust over time \citep{nadel2000multiple}. More recently, the transformation hypothesis extends this debate by proposing that consolidation concerns not only where memories are stored but also how their representational content changes over time \citep{winocur2010memory}. According to this view, initially context-rich episodic memories are gradually transformed into more abstract, decontextualized semantic or schematic representations. The hippocampus remains essential for detailed, context-dependent retrieval, whereas the neocortex supports the retention of gist-level knowledge. Crucially, this framework reconceptualizes retrieval as a dynamic reconstructive process shaped by stored information, current cues, task demands, and individual goals, rather than as a passive readout of fixed traces.

Despite their differences, these three theories share several fundamental premises. All acknowledge the hippocampus as critical for initial memory formation, all agree that memory is not static but undergoes change over time, and all recognize that reactivation plays an important role in this process. Where they diverge is in their account of what this change entails. SCT views consolidation as a spatial transfer from the hippocampus to the neocortex, after which the memory trace is fixed. MTT reframes it as the generation of multiple hippocampal traces that collectively strengthen the memory while preserving its dependence on the hippocampus. The transformation hypothesis departs from both by arguing that the change is fundamentally representational. What is consolidated is not the same memory relocated to a different region or replicated as multiple traces, but a qualitatively different, more schematic version of the original experience.

Among these perspectives, the transformation hypothesis resonates most closely with the computational challenges faced by modern sequence models. Its characterization of memory as an evolving, context-sensitive representation, rather than a fixed record to be passively retrieved, suggests a principled design direction for neural architectures that must maintain and adapt internal states over extended temporal horizons. At the same time, the neuroscience literature on system consolidation highlights that such memory transformation does not occur within a single structure in isolation, but emerges from the coordinated interaction of functionally distinct neural systems that operate at different temporal scales — a process mediated by cross-frequency coupling. These two insights jointly motivate the design of our model's memory module. Inspired by the principle of cross-frequency coupling, we propose a hierarchical memory module (HMM) composed of two independent sub-memory modules that coordinate with each other, with one operating at a high frequency and the other at a low frequency. This architecture incorporates both consolidation mechanisms into a unified framework. Synaptic consolidation finds a natural counterpart in standard gradient-based weight updates, which stabilize learned representations at the parameter level. The more substantial design challenge lies in realizing an analogue of system consolidation, a challenge that the proposed HMM architecture is designed to address. Furthermore, inspired by the transformation hypothesis, the final memory representations produced by HMM are not static outputs of any single sub-module but rather dynamic combinations of the representations generated by both. These combined representations are actively reconstructed and evolve at each forward step through interaction with incoming context. Because the two sub-memory modules differ in update frequency and model capacity, the representations they produce vary in their balance between episodic and semantic content, and their relative contribution to the final combination can be adaptively adjusted according to the input context.

\subsection*{Contributions}
In this paper, we aim to leverage the well-established consolidation theory and neuroscientific findings to build a novel hierarchical neural memory module that performs system-level memory consolidation at test time. Combined with a Transformer backbone, this hierarchical module enables a memory-augmented language model that constructs memory online during inference.

\textbf{HMM}. Motivated by cross-frequency coupling and the transformation hypothesis, we propose a Hierarchical Memory Module (HMM) comprising two sub-modules that differ in model capacity and the number of forward cycles. We also propose hierarchical latent recursion (HLR), a new algorithm that enables the two memory modules to interact in a manner analogous to HRM \citep{wang2025hierarchical} while circumventing the need for 1-step approximation. With HLR, the low-level memory module (L-module), which performs more forward cycles per pass and thus operates at a higher update frequency, produces low-level memory representations that retain richer episodic detail, whereas the high-level memory module (H-module), which performs fewer forward cycles per pass and operates at a lower update frequency, produces high-level memory representations that capture the gist of the input while discarding fine-grained contextual information. HMM embodies the three core principles of the transformation hypothesis. First, high-level memory remains dependent on the low-level memory. Second, during consolidation the low-level memory representation is not merely replicated into the high-level sub-module but undergoes transformation, so that the resulting high-level representation differs from the original. Third, the final output of HMM is a combination of the representations generated by the L-module and H-module, in which the contribution of each is determined by the query at retrieval time.

\textbf{Mela Architecture}. Treating the memory as a reference context, we connect HMM to a language decoder and present Mela, a family of models in three sizes that learns to memorize inputs and performs memory consolidation at test time. In contrast to interleaving memory operations within the decoder itself, Mela separates memory processing into a dedicated HMM that operates independently from the language decoder. This design echoes the modular organization of biological neural systems, where dedicated structures are specialized for distinct cognitive functions rather than being consolidated into a single unified network. Such separation yields practical benefits. The memory module can be independently scaled and trained with its own objective. Moreover, the modular design makes it straightforward to extend the system with new specialized modules requiring only minor modifications to the existing architecture. To better exploit the memory representations generated every cycle by the high-level memory module, we further propose MemStack, a novel method that stacks different levels of memory features across the top layers of the decoder. By leveraging MemStack, HMM provides fine-grained memory information to the decoder without introducing additional tokens, thereby preserving computational efficiency.

\textbf{Experiments}. We perform evaluation on the language modeling task to assess the performance of Mela. The Mela architecture outperforms the Transformer across all three model sizes, demonstrating the effectiveness of the proposed design. With the pretrained context length fixed at 4K, we compare the effective context lengths of Mela and Transformer and observe that Mela generalizes to significantly longer contexts than its pretrained window, whereas the Transformer's performance degrades rapidly beyond the training length. Finally, we conduct comprehensive ablation studies on various design choices of Mela, validating the contribution of each component in the hierarchical memory module and providing guidance for practical configuration.

\section{Preliminaries} \label{sec:preliminaries}
This section introduces the notation and technical preliminaries adopted throughout the paper. We denote $x \in \mathbb{R}^{N\times d}$ as the input, where $N$ is the sequence length and $d$ is the hidden dimension. We denote the neural memory module as a parameterized function 
$\mathcal{N}_{\mathcal{M}}: \mathbb{R}^{N \times d} \to \mathbb{R}^{N \times d}$, where $\mathcal{M}$ denotes its parameters. The subscript $t$ indicates the timestep.

\subsection{Neural Memory Module} 
Following the view that learning is the process of acquiring effective and useful memory, the neural memory module encodes such memory within its weight parameters, compressing historical information into them via gradient descent \citep{behrouz2024titans}. Under this formulation, the gradient serves as a measure of surprise, quantifying how much the input deviates from the model's current expectations. Specifically, given the input $x \in \mathbb{R}^{N\times d}$, the neural memory module updates the memory as:

\begin{equation} \label{eqn:basememory}
    \mathcal{M}_t = \mathcal{M}_{t-1} - \theta_t \nabla \ell(\mathcal{M}_{t-1}, x_t),
\end{equation}

where $\theta_{t}$ is a learnable learning rate at timestep $t$ and $\ell$ is an objective function. Compared to other modern recurrent variants that compress memory into fixed-size matrix-valued states \citep{katharopoulos2020transformers, lahoti2026mamba, yang2024gated, team2025kimi}, encoding memory into the weights of a neural network can yield more expressive memory representations, with the achievable expressiveness determined by the architecture and capacity of the underlying model. Beyond naive gradient descent, we can introduce momentum to prevent the model from stalling in flat regions that may arise after a sequence of highly surprising steps, and to accelerate learning when the update direction is consistent.

\begin{align}
    \mathcal{M}_t &= \mathcal{M}_{t-1} + S_t, \label{eqn:momentum_memory} \\
    S_t &= \eta_t \, S_{t-1} - \theta_t \, \nabla \ell(\mathcal{M}_{t-1}, x_t), \label{eqn:momentum_update}
\end{align}

where $\eta_t \in [0, 1]$ is a learnable decay factor that controls how much past surprise contributes to the weight update. In addition to the decay factor of surprise, we can also use the input-dependent forgetting factor frequently used in modern recurrent networks to filter the memory, which is empirically shown to be beneficial for retaining useful information under limited memory capacity \citep{gu2023mamba, yang2024gated}:

\begin{align}
    \mathcal{M}_t &= \alpha\mathcal{M}_{t-1} + S_t, \label{eqn:forget_momentum_memory} \\
    S_t &= \eta_t \, S_{t-1} - \theta_t \, \nabla \ell(\mathcal{M}_{t-1}, x_t), \label{eqn:forget_momentum_update}
\end{align}

where $\alpha \in [0, 1]$ is the forgetting factor adaptively regulating how much the past memory should retain at every step. When $\alpha \rightarrow 1$, the past memory will be kept entirely; when $\alpha \rightarrow 0$, the past memory will be fully forgotten. 

There are several design choices for the form of the objective function. The most naive choice is to train the neural memory module to reconstruct the original input $x$:

\begin{equation} \label{eqn:naive_loss}
    \ell(\mathcal{M}_{t-1}, x_t) = ||\mathcal{N}(x_t; \mathcal{M}_{t-1}) - x_{t}||_{2}^{2}.
\end{equation}

In principle, this objective encourages the memory module to faithfully reconstruct the input. However, this objective is suboptimal, as it incentivizes the model to focus on surface-level details rather than capturing underlying patterns that generalize across timesteps. As a result, the model may allocate substantial capacity to encoding redundant or noisy contexts, making it difficult to retain truly important information. Another choice is the associative loss, where the input is mapped into a key-value pair and the memory module learns the association between them. Concretely, given the weight matrices $W_{k}$ and $W_{v}$, the input $x_t$ is projected into the key and value as:

\begin{equation} \label{eqn:key_value_mapping}
    k_{t} = W_{k}x_{t},\quad v_{t} = W_{v}x_{t}, 
\end{equation}

and the associative loss is formulated as:

\begin{equation} \label{eqn:associative_loss}
    \ell({\mathcal{M}}_{t-1}, x_t) = ||\mathcal{N}(k_t; \mathcal{M}_{t-1}) - v_{t}||_{2}^{2}.
\end{equation}

From a memory perspective, what is learned by minimizing this objective is not a verbatim copy of the input, but the structured association between keys and values. This formulation encourages the memory module to capture relational information, specifically how different aspects of the input correspond to one another, rather than memorizing surface-level features. This aligns with the central thesis outlined earlier: the purpose of memory is not faithful reconstruction of past events, but the formation of structured references that support higher-level cognitive processes. Besides the associative $\ell_2$ loss function, general $\ell_p$-norm losses, Huber loss, elastic net, and other advanced loss functions are all viable alternatives for the memory objective \citep{behrouz2025s}. In this work, we primarily adopt the associative memory formulation due to its simplicity, computational efficiency, and well-established theoretical connection to key-value retrieval.

After storing information in memory, we use an input-dependent query vector to retrieve relevant content. Specifically, given the input $x_t$ and weight matrix $W_q$, we generate query $q_t = W_q x_t$ and retrieve the corresponding output $y_t$ from memory by:

\begin{equation} \label{eqn:retrieve_memory}
    y_t = \mathcal{N}(q_t; \mathcal{M}_{t-1}).
\end{equation}

\subsection{Deep Recursion}
Hierarchical latent recursion has recently been proposed as an alternative to Chain-of-Thought (CoT) \citep{wei2022chain} and Test-Time Compute (TTC) \citep{snell2024scaling} for enhancing the effective computation depth in reasoning tasks \citep{wang2025hierarchical}. Inspired by the hierarchical, multi-timescale coordination mechanisms across cortical regions during reasoning, the Hierarchical Reasoning Model (HRM) comprises two recurrent modules, $f_{L}$ (low-level) and $f_{H}$ (high-level), operating at different recurrent frequencies, where $f_{L}$ recurs at a higher frequency and $f_{H}$ at a lower frequency. Without pretraining or CoT, HRM has been shown to substantially outperform much larger models on challenging structured reasoning benchmarks that demand complex logical search, spatial planning, and abstract inductive reasoning \citep{wang2025hierarchical}. Given an input $x$, HRM performs $N$ high-level cycles, each consisting of $T$ low-level timesteps, yielding $NT$ total timesteps in a single forward pass. Throughout this process, the two recurrent modules maintain their own hidden states, $z_H$ for $f_H$ and $z_L$ for $f_L$. To avoid the prohibitive memory cost of backpropagation through the entire recurrence, HRM adopts a 1-step gradient approximation \citep{bai2019deep}: all timesteps except the final cycle are executed without tracking gradients, and the hidden states are detached before the last cycle is re-executed with gradients enabled. For instance, with $N{=}2$ cycles of $T{=}2$ timesteps, the forward process proceeds as follows:
\begin{align*}
z_L &\leftarrow f_L(z_L + z_H + x) \quad \text{\# without gradients} \\
z_L &\leftarrow f_L(z_L + z_H + x) \quad \text{\# without gradients} \\
z_H &\leftarrow f_H(z_L + z_H) \quad \text{\# without gradients} \\
z_L &\leftarrow f_L(z_L + z_H + x) \quad \text{\# without gradients} \\
z_L &\leftarrow z_L.\text{detach}() \\
z_H &\leftarrow z_H.\text{detach}() \\
z_L &\leftarrow f_L(z_L + z_H + x) \quad \text{\# with gradients} \\
z_H &\leftarrow f_H(z_L + z_H) \quad \text{\# with gradients}.
\end{align*}

After the recurrence completes, the final high-level hidden state $z_H$ is passed through an output network to generate the output. 

Although HRM achieves impressive performance on reasoning tasks, it relies on a strong assumption that the recurrence converges to a fixed point, so that the 1-step gradient approximation can be applied at that equilibrium. To bypass this assumption, Tiny Recursion Model (TRM) proposes deep supervision, which executes $T - 1$ recursion steps without tracking gradients before running the final recursion step with gradients enabled \citep{jolicoeur2025less}. To motivate a simpler architecture, the authors further argue that $z_H$ can be interpreted as the solution at the current cycle and $z_L$ as the latent reasoning state from which $z_H$ is derived. Under this perspective, TRM adopts a single network to implement the recurrence rather than a hierarchical architecture, and discards the biological interpretation underlying HRM. Under this design, TRM achieves higher accuracy than HRM on the same reasoning benchmarks. However, the authors of TRM also note that scaling up the number of layers fails to yield further improvements and instead leads to overfitting, suggesting a potential limitation in scalability. Our work draws on the latent recursion paradigm introduced by HRM and TRM, while reintroducing a neuroscience-grounded hierarchical design that addresses the scalability concerns observed in TRM.

\section{Hierarchical Memory Module: Neural Memory with Test-Time Consolidation}
\begin{figure}[t!]
  \centering
  \includegraphics[width=\textwidth]{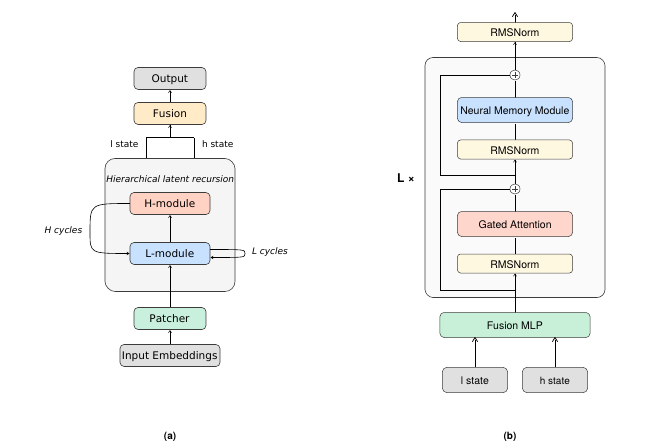}
  \caption{Overview of the Hierarchical Memory Module (HMM). (a)~Model architecture of Hierarchical Memory Module (HMM). At each forward pass, the input is initially patched to form episodic memory patches with fixed patch size. Then they are fed into the L-module and H-module to perform hierarchical latent recursion which can be seen as memory consolidation from a memory perspective. The output $h$ state and $l$ state are eventually combined through fusion layers. (b)~illustrates the architecture of memory modules, which consists of fusion MLP layer integrating the input $l$ state and $h$ state and a stack of memory blocks containing a gated attention followed by the neural memory module.}
  \label{fig:hmm_architecture}
\end{figure}

In this section we provide a detailed description of the proposed Hierarchical Memory Module (HMM) architecture. HMM generates two complementary forms of memory, low-level episodic memory and high-level semantic memory, from the input tokens, and performs memory consolidation through deep recursion at test time. The overall architecture of HMM is presented in Figure~\ref{fig:hmm_architecture}(a). In Section~\ref{subsec:memory_module}, we first introduce the memory module that serves as the shared backbone of both the H-module and the L-module for producing memory representations. Section~\ref{subsec:memory_consolidation} then describes how deep recursion coordinates the H-module and the L-module as an analogue of system consolidation, producing a high-level semantic memory state ($h$ state) and a low-level episodic memory state ($l$ state). Finally, inspired by the transformation hypothesis in memory neuroscience, Section~\ref{subsec:dual_memory_fusing} presents the fusing mechanism that integrates both memory traces into a unified output.

\subsection{Memory Module} \label{subsec:memory_module}
Figure~\ref{fig:hmm_architecture}(b) illustrates the overall architecture of the memory module. Input tokens are first projected into embeddings $x \in \mathbb{R}^{N \times d}$ through the embedding layer and then patchified into $x_p \in \mathbb{R}^{N_p \times d}$ by a 1-D convolution with a fixed patch size $P$, where $N$ denotes the input sequence length and $N_p = \lfloor N / P \rfloor$ the resulting number of patches. Each patch thus constitutes an episodic memory unit comprising $P$ consecutive tokens, which serves as the minimal unit entering the memory module. Following HRM, the initial $l$ state and $h$ state are both set to $x_p$ and fused through an MLP layer. Each memory block consists of a gated attention layer~\citep{qiu2025gated} followed by a neural memory module as introduced in Section~\ref{sec:preliminaries}, with RMSNorm~\citep{zhang2019root} applied prior to each sub-layer.

In the neural memory module, the patched sequence is further segmented into non-overlapping chunks of size $C$, so that the memory module processes $N_c = N_p / C$ chunks per forward pass. Specifically, within each chunk, the key and value projections are computed, and an associative loss is evaluated against the current memory state. The resulting gradient then serves as the surprise signal. The chunk size $C$ governs the temporal resolution of memory updates. Smaller values yield more frequent, fine-grained updates at higher computational cost, while larger values amortize computation but coarsen the update schedule. Moreover, we leverage Muon optimizer \citep{jordan2024muon}, which orthogonalizes momentum to accelerate and improve training and has demonstrated strong performance on large-scale LLM pretraining \citep{liu2025muon}. Specifically, the update rule of memory can be written as:

\begin{align}
    \mathcal{M}_t &= \alpha\mathcal{M}_{t-1} + \mathrm{NewtonShulz-}k(S_t), \label{eqn:newton_forget_momentum_memory} \\
    S_t &= \eta_t \, S_{t-1} - \theta_t \, \nabla \ell(\mathcal{M}_{t-1}, x_t), \label{eqn:newton_forget_momentum_update}
\end{align}
where $\mathrm{NewtonSchulz-}k(\cdot)$ is a k-steps iterative Newton-Schulz orthogonalization operation \citep{bjorck1971iterative, kovarik1970some}. As $k \rightarrow \infty$, $\text{NewtonShulz-}k(S_t)$ gradually converges to nearest semi-orthogonal matrix. Following \citet{jordan2024muon}, we set $k=5$, as five iterations have been shown to be sufficient for training.

Notably, the H-module and L-module are equipped with different numbers of memory blocks, endowing them with distinct model capacities. The H-module, which is responsible for generating abstract memory representations, is conceptually analogous to cortical regions in the human brain and accordingly comprises a greater number of memory blocks. Conversely, the L-module, which produces memory representations that retain richer episodic detail, parallels the role of the hippocampus and accordingly comprises fewer memory blocks.

\subsection{Memory Consolidation through Hierarchical Latent Recursion} \label{subsec:memory_consolidation}
\begin{figure}[t!]
  \centering
  \includegraphics[width=\textwidth]{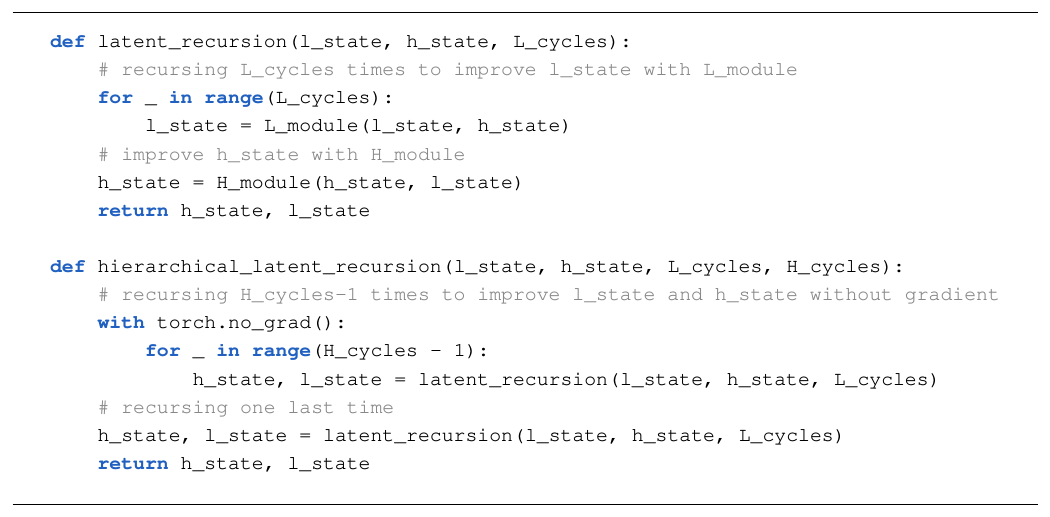}
  \caption{Pseudocode of HLR implemented in PyTorch.}
  \label{fig:hlr_pseudocode}
\end{figure}

Having constructed H-module and L-module from the memory module backbone, we employ the proposed hierarchical latent recursion (HLR) to jointly refine the $l$ state and $h$ state. The procedure is depicted in Figure~\ref{fig:hlr_pseudocode}. The recursion operates at two levels, namely a high-level recursion spanning $H$ cycles and a low-level recursion spanning $L$ cycles. Following the design of HRM, within each of the $H$ high-level cycles, the $l$ state undergoes $L$ refinement cycles through the L-module. Upon completion of these $L$ cycles, the H-module updates the $h$ state conditioned on the refined $l$ state. Because the $h$ state is updated only after the $l$ state has been refined $L$ times, the parameter $L$ effectively governs the update ratio between the two states. A larger value of $L$ thus implies a proportionally higher update frequency for the L-module relative to the H-module, and vice versa. Consistent with TRM, the $l$ state and $h$ state are refined $H-1$ times without gradient computation, followed by a single final refinement step in which gradients are propagated through the computation graph.

We interpret this entire recursion as a computational analogue of system consolidation within the transformation hypothesis framework. Under the transformation hypothesis, incoming memories in the human brain are initially encoded in both the hippocampus and neocortical regions, corresponding here to the $l$ state maintained by the L-module and the $h$ state maintained by the H-module, both initialized to the patched input $x_p$. This corresponds to cross-frequency coupling: the $h$ state and $l$ state  are updated at different frequencies determined by the ratio $L$, and through their interaction, memory representations undergo progressive transformation during consolidation.

\subsection{Dual Memory Fusion} \label{subsec:dual_memory_fusing}
Once the last recursion completes, the $h$ state and the $l$ state are integrated to produce the final output. Several methods exist for generating the final output from the $h$ state and $l$ state. One naive choice is to use only the $h$ state and discard $l$ state as HRM and TRM do. In our implementation, we concatenate the $h$ state and $l$ state and feed them into a 2-layer MLP to generate the final output.

Overall, the advantages of HMM are three-fold. First, through HLR, we construct a refinement pipeline that efficiently improves the $l$ state and $h$ state without relying on the fixed-point convergence assumption required by HRM. Second, unlike HRM and TRM, HMM combines the $l$ state and $h$ state in its final output, ensuring that the representation retains both high-level and low-level information. Third, HMM establishes a connection to neuroscience that extends beyond HRM and is absent in TRM. Specifically, the interplay between the $l$ state and the $h$ state mirrors the memory transformation process described by the transformation hypothesis. While we do not claim functional equivalence with biological memory systems, we argue that grounding architectural choices in well-studied neuroscience principles offers a principled design framework, as these principles reflect mechanisms that support complex cognition in the only known system capable of reasoning, planning, and introspection.

\section{Mela Architecture}
\begin{figure}[t!]
  \centering
  \includegraphics[width=\textwidth]{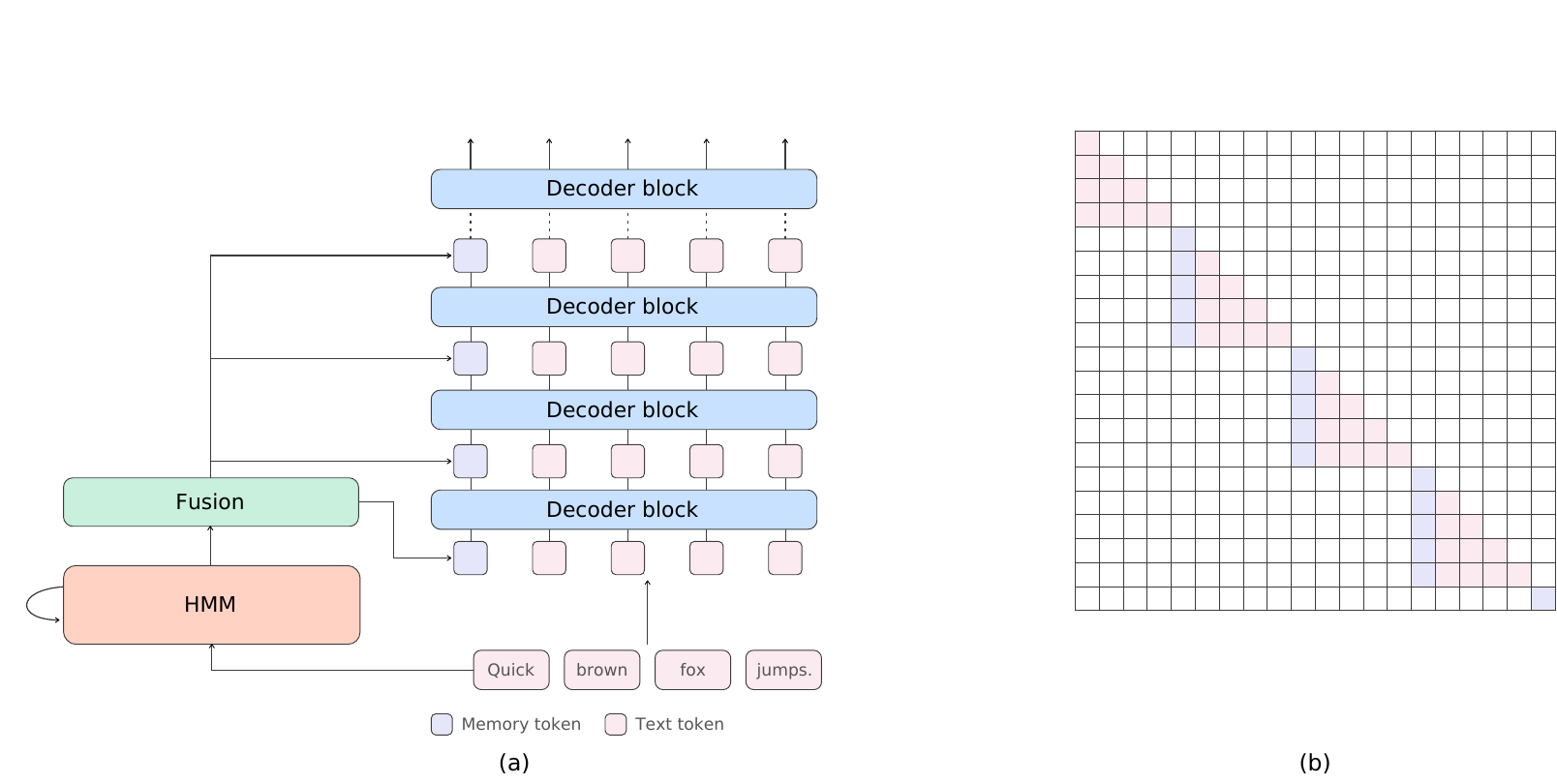}
  \caption{(a)~Model architecture of Mela. During the forward pass, HMM recurs over $H$ cycles to produce the $l$ states and $h$ states at each cycle. In addition to feeding the memory features from the last cycle into the input layer of the language decoder, the intermediate $l$ states and $h$ states from each earlier cycle also pass through their own fusion layers to produce intermediate memory features. These intermediate memory features are then infused into early decoder layers. This strategy, denoted as MemStack, allows the model to progressively acquire memory information from all cycles without introducing extra memory tokens. (b)~illustrates an example attention mask of Mela with patch size 4 and chunk size 1. The input sequence is segmented and the memory tokens are prepended as reference contexts.}
  \label{fig:mela_and_attention_mask}
\end{figure}

In this section, we present a detailed description of the Mela architecture, which leverages HMM to produce memory representations that serve as reference contexts for the language decoder. Figure~\ref{fig:mela_and_attention_mask}(a) shows the overall architecture of Mela. In Section~\ref{subsec:incorporate_memory}, we introduce how to incorporate memory into the decoder. In Section~\ref{subsec:memstack}, we further describe the proposed MemStack method, which infuses memory representations from each of the $H$ recurrence cycles into the early decoder layers, allowing the decoder to leverage the full consolidation trajectory rather than only the final memory state. 

\subsection{Incorporating Memory into the Input Context} \label{subsec:incorporate_memory}
After HMM outputs the memory representations through HLR, the next question is how to effectively utilize these memories for the downstream task. In this work, we focus on language modeling, where the memory representations are derived from the text tokens themselves. Given embedded input $x \in \mathbb{R}^{N \times d}$, HMM produces memory representations $m \in \mathbb{R}^{N_p \times d}$.

To incorporate these memories into the language decoder, we treat them as reference contexts. Specifically, we segment the input text tokens into chunks of size $PC$ and prepend $C$ memory features to each chunk. For a sequence of length $L$ with patch size $P$ and chunk size $C$, this yields $L / PC$ chunks and a total input length of $L(P+1)/P$. Figure~\ref{fig:mela_and_attention_mask}(b) illustrates an example of the resulting attention mask with patch size 4 and chunk size 1. Under this design, each text token attends only to the memory tokens and preceding text tokens within its own chunk, with no direct attention across chunk boundaries. Cross-chunk information is therefore carried entirely by the prepended memory features, which compress all preceding context into a fixed number of memory tokens. This makes HMM the sole information channel between chunks, ensuring that the language decoder is forced to rely on the memory representations rather than fall back on direct attention to past tokens.

\subsection{MemStack} \label{subsec:memstack}
During HLR, the H-module and L-module produce a sequence of $l$ states and $h$ states, one pair per cycle. These can be viewed as intermediate low-level and high-level memory states at increasing consolidation depths. Compared to using only the final output of HMM as the source of past information, we pass the intermediate $l$ states and $h$ states from each cycle through their corresponding fusion layers to produce intermediate memory representations that capture the consolidation state at each cycle. Among various fusion strategies, stacking features across decoder layers has been shown to be an effective way to integrate information without introducing extra computational cost \citep{meng2024deepstack, bai2025qwen3}. Formally, let $l^{(i)}$ and $h^{(i)}$ denote the $l$ state and $h$ state generated at cycle $i$, where $i \in \{1,\ldots, H\}$. We form the memory representation at each cycle through the fusion layer:

\begin{equation} \label{eqn:fusion_mlp}
    z^{(i)} = \mathrm{Fusion}(l^{(i)}, h^{(i)}),
\end{equation}
where $z^{(i)}$ is the memory representation generated at cycle $i$ and $\mathrm{Fusion}(\cdot)$ is the fusing operation. The last cycle produces the final memory representation $z^{(H)}$, while the earlier cycles produce intermediate memory representations $z^{(j)}$ for $j \in \{1,\ldots,H-1\}$. Let $B^{(\alpha)}$ denote the $\alpha$-th decoder block, where $\alpha \in \{1, \ldots, L\}$ and $L$ is the total number of decoder layers. MemStack injects the intermediate memory representations $z^{(j)}$ into the first $H-1$ decoder layers. Denote the output hidden state after the $\alpha$-th decoder layer as $y^{(\alpha)}$ where $\alpha \in \{1,\ldots, L\}$, MemStack can be expressed as follows:

\begin{equation} \label{eqn:mem_stack_formula}
\begin{aligned}
    y^{(1)} &= B^{(1)}(z^{(H)}) + z^{(1)} \\
    y^{(2)} &= B^{(2)}(y^{(1)}) + z^{(2)} \\
    ... \\
    y^{(H-1)} &= B^{(H-1)}(y^{(H-2)}) + z^{(H-1)} \\
    y^{(H)} &= B^{(H)}(y^{(H-1)}) \\
    ... \\
    y^{(L)} &= B^{(L)}(y^{(L-1)}).
\end{aligned}
\end{equation}

For brevity, $y^{(\alpha)}$ refers to the hidden state at memory token positions; text token positions follow the standard decoder forward pass without modification.Interpreting $z^{(i)}$ as intermediate outputs of consolidation, the memory tokens carry not only the final consolidated result but also information about the consolidation process itself. The forward pass of memory features inside the language decoder is thus divided into two phases. In the early layers, the decoder encodes both the memory content and the consolidation trajectory, enriching the memory representation. The later layers operate as standard decoder layers for sequence modeling. The full forward pass of the language decoder can be written more concisely as:

\begin{equation} \label{eqn:concise_memstack_formula}
  y^{(L)} = B^{(L)}\left(\dotsb \left(B^{(H-1)}\left(\dotsb B^{(1)}\left(z^{(H)}\right) + z^{(1)}\dotsb\right) + z^{(H-1)}\right)\dotsb\right)
\end{equation}

This design parallels the transformation hypothesis: rather than presenting only the fully consolidated memory to downstream processing, MemStack exposes the decoder to memory representations at multiple stages of consolidation, mirroring how the brain may access memory traces at different levels of abstraction during cognition.

\section{Experiments}
We evaluate the performance of the proposed Mela architecture across different sizes on language modeling task. We show that Mela outperforms baseline and achieves extended context window size with fixed pre-trained context length. In this section, we also performs ablation studies to verify the effectiveness of key designs in Mela and check how do they impact the effective context length in pratical.

\subsection{Experimental Setup} \label{subsec:experimental_setup}

\begin{table}[t]
\centering
\setlength{\tabcolsep}{12pt}
\renewcommand{\arraystretch}{1.2}
\begin{tabular}{@{}lccc@{}}
\toprule
\textbf{Hyperparameter} & \textbf{Mela-400M} & \textbf{Mela-800M} & \textbf{Mela-1.2B} \\
\midrule
H cycles ($H$)                 & 4     & 4     & 4     \\
L cycles ($L$)                 & 4     & 4     & 4     \\
H-module layers                 & 8     & 8     & 8     \\
L-module layers                 & 4     & 8     & 8     \\
HMM hidden size        & 768     & 1024     & 1024     \\
Decoder hidden size        & 768     & 1024     & 1536     \\
Patch size               & 32    & 32    & 32    \\
Chunk size               & 64    & 64    & 64    \\
Memory proportion (\%)   & 27.45 & 28.78 & 18.19 \\
Language proportion (\%) & 65.75 & 66.35 & 76.97 \\
\bottomrule
\end{tabular}
\caption{Hyperparameters for Mela variants. The remaining parameters correspond to embedding and output projection layers.}
\label{tab:mela-hparams}
\end{table}

\textbf{Models} To examine the scaling behavior of the proposed Mela architecture, we evaluate models at three scales, namely 400M, 800M, and 1.2B parameters. The detailed configurations for each scale are reported in Table~\ref{tab:mela-hparams}. All variants employ HLR to produce memory representations, with the H cycle and the L cycle both set to 4. We use a patch size of 32 and a chunk size of 64 for all variants. For the 800M variant, we increase the depth of the L-module to strengthen the quality of the $l$ state. For the 1.2B variant, we instead allocate the additional parameters to the language decoder by increasing its hidden size. Memory-related parameters account for roughly 30$\%$ of the total parameters in the 400M and 800M variants. For the 1.2B variant, we shift this allocation toward the language decoder (raising its share to 77$\%$) in order to enhance encoding performance. Across all variants, memory is updated using the Newton–Schulz method given in Equation~\ref{eqn:newton_forget_momentum_memory} and Equation~\ref{eqn:newton_forget_momentum_update}.

\textbf{Baselines} We compare against a modern Transformer baseline matched to Mela at each of the three parameter scales, which incorporates several advanced refinements over the original implementation \citep{vaswani2017attention}, including RMSNorm \citep{zhang2019root}, SwiGLU \citep{shazeer2020glu, ramachandran2017searching}, Rotary Position Embedding (RoPE) \citep{su2021roformer}, and grouped-query attention (GQA) \citep{ainslie2023gqa}. This configuration shares the same architecture as LLaMA \citep{touvron2023llama} and is commonly referred to as Transformer++. Both Transformer++ and Mela use Llama 2 tokenizer \citep{touvron2023llama2} with a vocabulary size of 32K.

\textbf{Training} The training dataset consists of 5B text tokens sampled from the FineWeb-Edu dataset \citep{penedo2024fineweb}, and the evaluation set is also drawn from this dataset. The training context length is set to 4K. All models are trained using the AdamW optimizer \citep{loshchilov2017decoupled}, with $\beta_1 = 0.9$ and $\beta_2 = 0.95$. For 400M and 800M model, the peak learning rate is set to $10^{-3}$. For the 1.2B models, the peak learning rate is set to $8\times 10^{-4}$. We use a cosine learning rate schedule with the minimum learning rate set to $10\%$ of the peak. We use warmup steps of 500, weight decay of 0.1, and gradient clipping of 1.0. The effective batch size is 0.5M tokens. For efficient distributed training, we use FSDP2 (the successor of FSDP \citep{zhao2023pytorch}) combined with gradient accumulation.

\subsection{Quantiative Results} \label{subsec:quantitive_results}

\begin{figure}[t!]
  \centering
  \includegraphics[width=\textwidth]{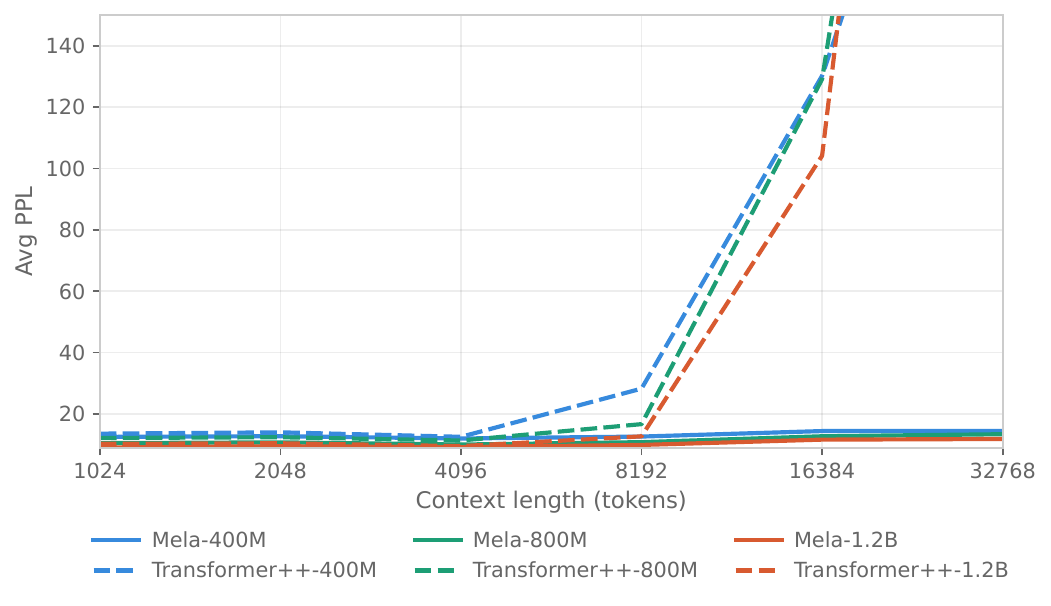}
  \caption{Perplexity of Mela models and Transformer++ baselines on the test dataset across different context lengths. All models are pre-trained with context length 4096.}
  \label{fig:comparison_between_mela_and_transformer}
\end{figure}

We evaluate the perplexity of Mela and the baseline in language modeling across three model sizes and multiple context lengths. Figure~\ref{fig:comparison_between_mela_and_transformer} visualizes the results, and Table~\ref{tab:ppl_results} reports the detailed numbers. At matched model size, Mela outperforms the baseline at the 4K pre-trained context length and maintains its advantage at both shorter and longer contexts. Beyond the pre-trained context length, the gap widens substantially. Without any additional length-extrapolation technique, the Transformer baseline's perplexity rises sharply once the context exceeds 4K and reaches an order of magnitude higher at 32K. In contrast, Mela exhibits only a mild increase across the extended range, with its 32K perplexity remaining on the same order of magnitude as its 4K result. We attribute this to the design of HMM and its integration with the language decoder. When the inference context surpasses the pre-trained length, HMM supplies the decoder with memory representations that combine episodic detail (from the L-module) and semantic abstraction (from the H-module), enabling reliable next-token prediction.

\subsection{Ablation Studies} \label{subsec:ablation_studies}
In this section, we aim to answer the following questions: (1) How does the recursion depth of HMM (i.e., the number of H cycles and L cycles) affect the performance of Mela across different context lengths? (2) How does the capacity of the H-module and L-module influence Mela's performance across different context lengths? (3) What is the contribution of each core component in the Mela architecture? To answer these questions, we conduct ablation studies on Mela models with approximately 400M parameters, which we treat as the default scale throughout this section unless stated otherwise. All variants are trained on a 1B-token subset uniformly sampled from the FineWeb-Edu dataset~\citep{penedo2024fineweb}, so that observed differences across configurations primarily reflect architectural choices rather than disparities in training data. For evaluation, we use a held-out split of FineWeb-Edu and report perplexity across multiple context lengths, which allows us to probe how each ablation affects both short- and long-context behavior under a consistent data distribution.

\begin{figure}[H]
  \centering
  \includegraphics[width=0.8\textwidth]{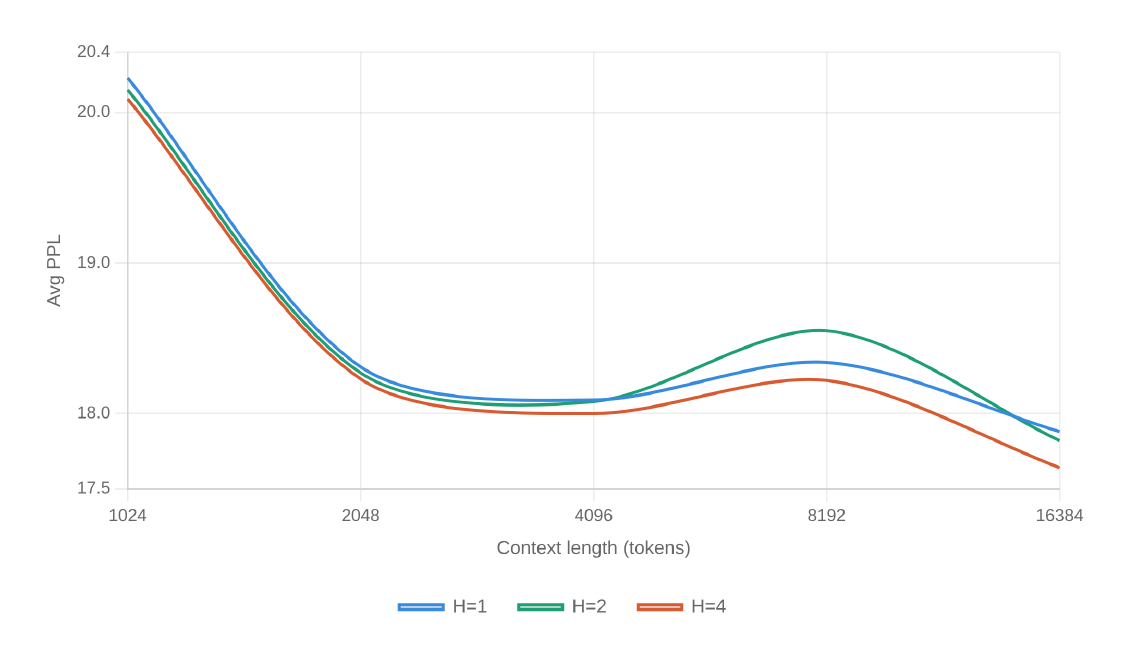}
  \caption{Effect of the number of $H$ cycles on Mela's perplexity across context lengths. Increasing $H$ from 1 to 4 consistently lowers perplexity, with the largest gains observed at long contexts beyond the 4K pre-training window. All variants share the 400M default configuration apart from the $H$ cycle count.}
  \label{fig:ablation_Hcycle_ablation}
\end{figure}

\begin{figure}[H]
  \centering
  \includegraphics[width=0.8\textwidth]{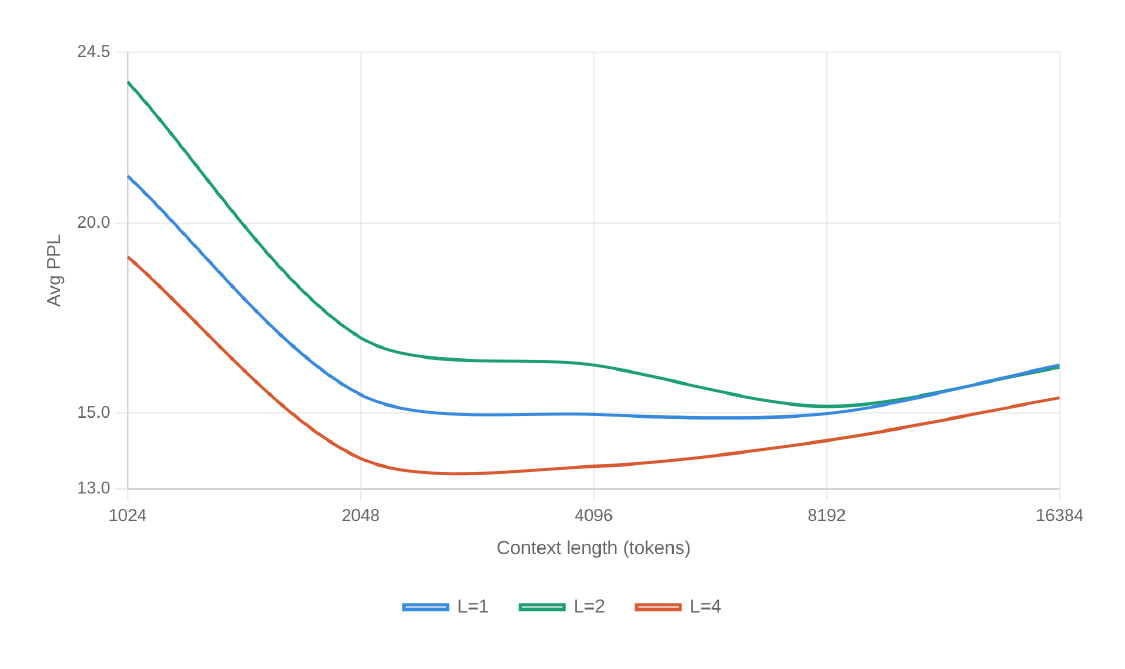}
  \caption{Effect of the number of $L$ cycles on Mela's perplexity across context lengths. $L=4$ achieves the lowest perplexity at all context lengths, while the intermediate setting $L=2$ underperforms $L=1$ at most lengths, suggesting that two L-module steps are insufficient to produce a meaningfully refined episodic representation. All variants share the 400M default configuration apart from the $L$ cycle count.}
  \label{fig:ablation_Lcycle_ablation}
\end{figure}

\textbf{Recurrent Forward Depth} We first examine how the recurrent forward depth of HMM affects Mela's performance. We sweep both the H cycle and the L cycle over the values \{1, 2, 4\}, holding all other architectural choices fixed at the 400M default, and report perplexity across context lengths in Table~\ref{tab:cycle_context}. The H cycle and L cycle portions of the table are visualized in Figure~\ref{fig:ablation_Hcycle_ablation} and Figure~\ref{fig:ablation_Lcycle_ablation}.

In principle, a larger number of $H$ cycles corresponds to a deeper recurrent forward depth and, intuitively, a more thorough consolidation process. The $H$ cycle portion of Table~\ref{tab:cycle_context} supports this expectation, as $H=4$ achieves the lowest perplexity in every column, with the trend approximately monotonic in $H$ for most context lengths. The improvement holds uniformly across short and long contexts, indicating that additional $H$ cycles translate into consistent gains regardless of sequence length. Notably, the gap between $H=2$ and $H=4$ expands from 0.06 at 1K tokens to 0.18 at 16K tokens, indicating that the benefit of additional consolidation grows at longer contexts. This pattern suggests that deeper recurrent updates become increasingly beneficial as the model is required to consolidate information over longer horizons, and it supports the role of $H$ cycles as the central consolidation mechanism in HMM.

We next examine the impact of the number of L cycles, which controls the forward frequency between the L-module and the H-module. Concretely, an L cycle of $k$ means that the L-module is unrolled for $k$ steps before each H-module forward pass, so the H-module consumes a memory representation that has been progressively refined by the L-module. A larger L cycle therefore supplies the H-module with a more thoroughly processed episodic memory representation, which we expect to translate into higher-quality consolidated outputs. As shown in the L cycle portion of Table~\ref{tab:cycle_context}, L=4 yields the best perplexity across all context lengths and improves substantially over L=1, for example from 21.24 to 19.11 at 1K tokens and from 14.96 to 13.59 at 4K tokens, supporting the view that a more refined L-module representation benefits downstream consolidation. The intermediate setting L=2 is an exception, as it underperforms L=1 at several context lengths and rises from 21.24 to 23.72 at 1K tokens, suggesting that two L-module steps are insufficient to produce a meaningfully refined representation while still incurring the cost of delayed H-module updates. Interestingly, the L cycle exhibits the opposite trend to the H cycle along the context-length axis, as the gap between L=2 and L=4 narrows from 4.61 at 1K tokens to 0.80 at 16K tokens. We interpret this as evidence that a coarsely refined episodic representation is most damaging at short contexts, where each token contributes a larger fraction of the available signal, whereas longer contexts provide enough redundancy for the H-module to partially compensate for an under-refined L-module representation.

\begin{table}[t]
\centering
\small
\setlength{\tabcolsep}{5pt}
\renewcommand{\arraystretch}{1.15}
\definecolor{gapgray}{gray}{0.55}
\newcommand{\gap}[1]{{\textcolor{gapgray}{\scriptsize\,(+#1)}}}
\newcommand{\best}[1]{\textbf{#1}}

\begin{tabular}{c|ccccc}
\toprule
\textbf{Context length} & \textbf{1024} & \textbf{2048} & \textbf{4096} & \textbf{8192} & \textbf{16384} \\
\midrule
\multicolumn{6}{c}{\textbf{H cycle}} \\
\midrule
H=1 & 20.23\gap{0.14} & 18.31\gap{0.08} & 18.09\gap{0.09} & 18.34\gap{0.12} & 17.88\gap{0.24} \\
H=2 & 20.15\gap{0.06} & 18.27\gap{0.04} & 18.08\gap{0.08} & 18.55\gap{0.33} & 17.82\gap{0.18} \\
H=4 & \best{20.09}    & \best{18.23}    & \best{18.00}    & \best{18.22}    & \best{17.64} \\
\midrule
\multicolumn{6}{c}{\textbf{L cycle}} \\
\midrule
L=1 & 21.24\gap{2.13} & 15.48\gap{1.68} & 14.96\gap{1.37} & 14.98\gap{0.71} & 16.26\gap{0.86} \\
L=2 & 23.72\gap{4.61} & 16.98\gap{3.18} & 16.26\gap{2.67} & 15.17\gap{0.90} & 16.20\gap{0.80} \\
L=4 & \best{19.11}    & \best{13.80}    & \best{13.59}    & \best{14.27}    & \best{15.40} \\
\bottomrule
\end{tabular}
\caption{Perplexity across context lengths for cycle ablations. Best per column (within each group) is \textbf{bolded}, with the gap to the best shown in parentheses.}
\label{tab:cycle_context}
\end{table}

\begin{figure}[H]
  \centering
  \includegraphics[width=0.8\textwidth]{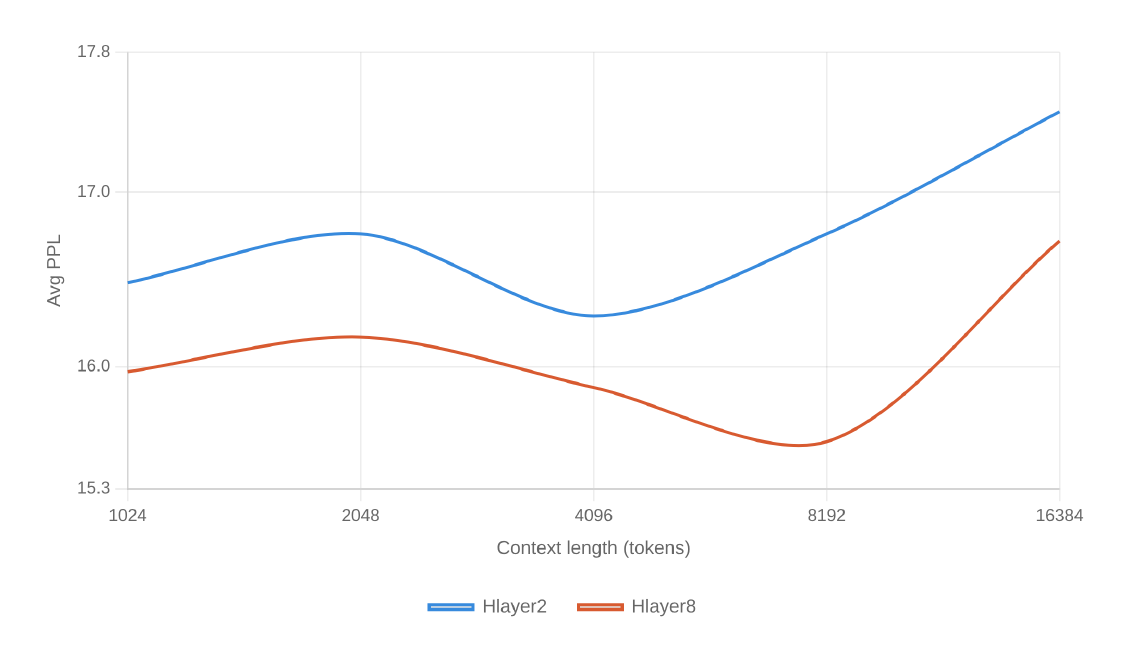}
  \caption{Effect of H-module depth on Mela's perplexity across context lengths. Increasing the H-module from 2 to 8 memory blocks improves perplexity at every context length, with the gap widening beyond the 4K pre-training window. All variants share the 400M default configuration apart from the H-module depth.}
  \label{fig:ablation_Hlayer_ablation}
\end{figure}

\begin{figure}[H]
  \centering
  \includegraphics[width=0.8\textwidth]{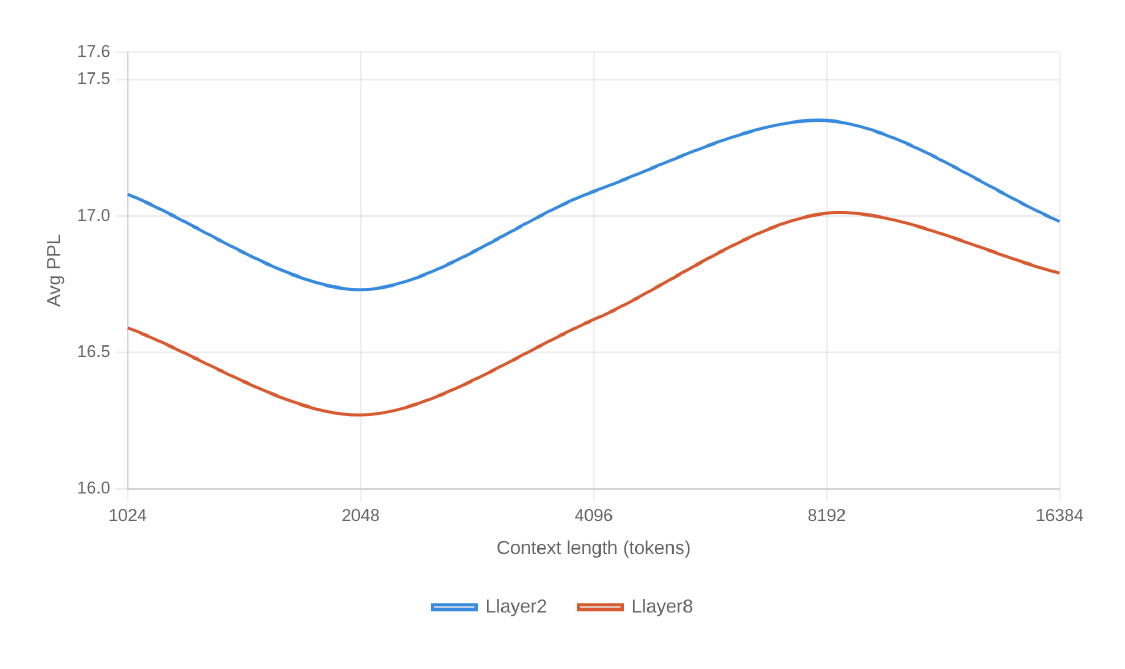}
  \caption{Effect of L-module depth on Mela's perplexity across context lengths. Increasing the L-module from 2 to 8 memory blocks improves perplexity at every context length, with the largest gains at short contexts. All variants share the 400M default configuration apart from the L-module depth.}
  \label{fig:ablation_Llayer_ablation}
\end{figure}

\textbf{HMM Module Capacity} The number of memory blocks in each module directly determines the representational capacity of the H-module and the L-module, and therefore shapes how effectively each module encodes memory. To isolate the contribution of capacity, we compare two depth settings (2 and 8 memory blocks) for each module independently and report the resulting perplexities in Table~\ref{tab:layer_context}, with the corresponding visualizations provided in Figure~\ref{fig:ablation_Hlayer_ablation} and Figure~\ref{fig:ablation_Llayer_ablation}. Across all context lengths, both modules benefit from the additional capacity afforded by deeper stacking, confirming that increased depth in either way translates into improved memory encoding within HMM. Beyond this overall trend, however, the two modules respond to additional capacity in markedly different ways along the context-length axis. For the L-module, the benefit of stacking is most pronounced at short contexts and gradually diminishes as the context grows, with the gap between layer=2 and layer=8 shrinking from 0.49 at 1K tokens to 0.19 at 16K tokens. This pattern is consistent with the L-module's role as an episodic encoder. At short contexts, the H-module has accumulated few consolidated traces and is therefore highly sensitive to the quality of the L-module's output. As the context lengthens, the H-module gradually builds up enough consolidated information to partially compensate for a less capable L-module, reducing the marginal benefit of additional L-module capacity. The H-module exhibits the opposite trend, as the gap between layer=2 and layer=8 generally widens with context length and reaches 1.19 at 8K tokens and 0.74 at 16K tokens, both substantially larger than the 0.51 gap observed at 1K tokens. We attribute this divergence to the consolidation role of the H-module, namely that integrating episodic traces into a coherent semantic representation becomes more challenging as the number of traces grows, so additional capacity yields larger returns in the long-context regime.

\begin{table}[t]
\centering
\small
\setlength{\tabcolsep}{5pt}
\renewcommand{\arraystretch}{1.15}
\definecolor{gapgray}{gray}{0.55}
\newcommand{\gap}[1]{{\textcolor{gapgray}{\scriptsize\,(+#1)}}}
\newcommand{\best}[1]{\textbf{#1}}

\begin{tabular}{c|ccccc}
\toprule
\textbf{Context length} & \textbf{1024} & \textbf{2048} & \textbf{4096} & \textbf{8192} & \textbf{16384} \\
\midrule
\multicolumn{6}{c}{\textbf{L-module}} \\
\midrule
layer=2 & 17.08\gap{0.49} & 16.73\gap{0.46} & 17.09\gap{0.47} & 17.35\gap{0.34} & 16.98\gap{0.19} \\
layer=8 & \best{16.59}    & \best{16.27}    & \best{16.62}    & \best{17.01}    & \best{16.79} \\
\midrule
\multicolumn{6}{c}{\textbf{H-module}} \\
\midrule
layer=2 & 16.48\gap{0.51} & 16.76\gap{0.59} & 16.29\gap{0.41} & 16.76\gap{1.19} & 17.46\gap{0.74} \\
layer=8 & \best{15.97}    & \best{16.17}    & \best{15.88}    & \best{15.57}    & \best{16.72} \\
\bottomrule
\end{tabular}
\caption{Perplexity across context lengths for layer ablations. Best per column (within each group) is \textbf{bolded}, with the gap to the best shown in parentheses.}
\label{tab:layer_context}
\end{table}

\begin{table}[t]
\centering

\begin{tabular}{llc}
\toprule
\textbf{Component} & \textbf{Variant} & \textbf{PPL} $\downarrow$ \\
\midrule
\multirow{2}{*}{Spectral Norm}     & w/o Newton-Schulz  & 16.30 (+0.21) \\
                                    & \textbf{w/ Newton-Schulz}   & \textbf{16.09} \\
\midrule
\multirow{2}{*}{MemStack}          & w/o MemStack       & 16.40 (+0.1) \\
                                    & \textbf{w/ MemStack}        & \textbf{16.30} \\
\midrule
\multirow{2}{*}{Gated Attention}  & w/o Gating        & 17.64 (+1.24) \\
                                    & \textbf{w/ Gating}          & \textbf{16.40} \\
\midrule
\multirow{3}{*}{Fusion Implementation}      & Weighted Sum       & 16.68 (+0.38) \\
                                    & None               & 16.41 (+0.11) \\
                                    & \textbf{MLP}                & \textbf{16.30} \\
\midrule
\multirow{2}{*}{Attention in Memory module}  & w/o Attention & 16.43 (+0.13) \\
                                    & \textbf{w/ Attention} & \textbf{16.30} \\
\midrule
\multirow{2}{*}{Neural Memory}     & w/o Neural Memory  & 16.73 (+0.43) \\
                                    & \textbf{w/ Neural Memory}   & \textbf{16.30} \\
\bottomrule
\end{tabular}

\caption{Ablation study results measured by perplexity (PPL), lower is better. The \textbf{bold} row indicates the full model configuration.}
\label{tab:additional_ablations}
\end{table}

\textbf{Consolidation and Long-Term Memory Formation} Taken together, the cycle and layer ablations reveal a coherent pattern across the H-module and the L-module when examined along the context-length axis. On the cycle side, deeper recurrent updates in the H-module deliver gains that become substantially larger once the context exceeds the 4K-token pre-training window, with the gap between $H=2$ and $H=4$ remaining modest within the pre-training window (0.06–0.08 at 1K–4K tokens) but expanding beyond it (0.33 at 8K tokens, 0.18 at 16K tokens). On the layer side, the same asymmetry appears with respect to module capacity, as additional H-module layers contribute markedly more once the sequence exceeds the pre-training window and reach a gap of 1.19 at 8K and 0.74 at 16K tokens, while additional L-module layers yield the largest gains at short contexts and provide diminishing returns as the context lengthens. The two ablation axes therefore converge on the same qualitative trend, namely that the H-module acts as the primary performance bottleneck at long contexts and the L-module acts as the primary bottleneck at short contexts. Viewed through the lens of the transformation hypothesis, these results admit a unified interpretation. The consolidation process implemented by the H-module appears to improve the quality of the final output memory representation in a manner that scales with the temporal horizon, which is consistent with the role of consolidation in supporting long-term memory formation. The L-module, by contrast, primarily contributes high-quality episodic information whose marginal value is largest when the context is short and the model has fewer episodic traces to draw on. A natural reading of the combined evidence is that effective use of episodic content is the limiting factor at short contexts, while effective consolidation into a coherent semantic representation becomes the limiting factor as the sequence grows beyond the pre-training window.

\textbf{Additional Ablations} Beyond the cycle and capacity studies, we further ablate several core design choices in Mela and report the resulting perplexities in Table~\ref{tab:additional_ablations}. We first examine two design choices that affect how memory information is learned and propagated. Replacing Newton-Schulz orthogonalization with its non-orthogonalized counterpart raises perplexity from 16.09 to 16.30, which suggests that approximately orthogonal weight updates stabilize the learning of memory parameters and prevent the kind of update interference that tends to accumulate when the H-module and the L-module are trained jointly over many recurrent steps. Removing MemStack produces a smaller but consistent degradation, and although the absolute gap of 0.10 is modest at the perplexity level, the result is consistent with our motivation for MemStack, namely that exposing the model to intermediate states of the consolidation process, rather than only the final episodic and semantic memories, supplies the decoder with additional information about the consolidation trajectory itself, which would otherwise be discarded once consolidation completes.

We next study two design choices that govern the expressivity of the memory module. The largest single effect among these components arises from the gating mechanism applied to the attention module within HMM, whose removal increases perplexity by 1.24 points and yields the worst configuration in this ablation table. We attribute this large gap to the selective-write role that gating plays in the memory module, since the H-module and the L-module operate on signals at very different temporal scales and the gate allows the model to suppress irrelevant traces before they are written into the consolidated representation. Without this non-linearity, every incoming signal contributes uniformly to the memory update and the resulting representation becomes harder to disentangle. A related set of ablations on how the $l$ state and the $h$ state are combined reinforces this view. A two-layer MLP achieves the best perplexity at 16.30, the H-state-only baseline reaches 16.41, and a learned weighted summation performs worst at 16.68 despite introducing additional parameters. The fact that weighted summation underperforms even the parameter-free H-state-only baseline is, on its face, counterintuitive, but it admits a natural reading under the H/L decomposition, namely that a single set of mixing weights forces the model to commit to a fixed trade-off between the two states across all token positions, whereas the relative usefulness of episodic and semantic information varies substantially with position. An MLP recovers performance because it can implement an input-dependent combination, while a linear mixer adds noise without adding flexibility.

Finally, we ablate the attention module and the neural memory module that operate inside the H-module and the L-module. Both components contribute to the final performance, but the neural memory contributes substantially more, with its removal raising perplexity by 0.43 against 0.13 for the attention module. We interpret this asymmetry through the lens of the transformation hypothesis. Attention provides flexible in-context retrieval but operates over a bounded window and does not, by itself, retain information across consolidation steps, whereas the neural memory acts as a persistent store whose state is updated and read across the full recurrent depth of HMM. The two mechanisms are therefore complementary rather than redundant, with attention supporting fine-grained access within a local window and the neural memory supporting the cross-segment integration that the transformation hypothesis posits as the substrate of long-term memory formation. The ablation results are consistent with this division of labor, since removing the component that supports cross-segment integration produces a noticeably larger degradation than removing the component that supports local access.

\section{Conclusion}
In this paper, we present the HMM architecture to generate memory representations at test time through a neural memory module, and propose the HLR mechanism inspired by cross-frequency coupling and the transformation hypothesis. Combining HMM with the language decoder, we present the Mela architecture and propose MemStack, a strategy that enriches memory representations in the decoder by exposing it to both the final consolidation output and the intermediate consolidation states. Through evaluations across different sizes, we show that under the same pre-trained context length, Mela outperforms a strong Transformer baseline at all scales and generalizes effectively beyond the pre-trained context length. These results suggest that Mela offers an effective alternative for extending context length in settings where compute or long-context training data are limited.

We believe that memory plays a foundational role in many cognitive functions that current models do not yet capture well, ranging from long-horizon planning to more open-ended capabilities. Neural memory appears to be a promising direction for endowing models with such capabilities, and may contribute to progress toward more general and capable forms of intelligence. To support further progress in this direction, we release the full implementation of Mela, and we hope that our work encourages continued exploration of memory-augmented architectures and test-time memory formation as part of the broader effort to build the next generation of AI models.

\clearpage

\bibliographystyle{plainnat}
\bibliography{references}

\clearpage
\beginappendix

\section{Detailed Perplexity Results by Context Length}
\begin{table}[h]
\centering

\begin{tabular}{llrrrrrr}
\toprule
\textbf{Model} & \textbf{Params} & \textbf{1024} & \textbf{2048} & \textbf{4096} & \textbf{8192} & \textbf{16384} & \textbf{32768} \\
\midrule
Transformer++ & 400M  & 13.59 & 14.02 & 12.56 & 28.26 & 130.21 & 303.56 \\
Transformer++ & 800M  & 12.17 & 12.42 & 11.35 & 16.69 & 129.10 & 497.85 \\
Transformer++ & 1.2B  & 10.27 & 10.46 &  9.53 & 12.71 & 104.14 & 597.37 \\
\midrule
Mela(ours)        & 400M  & 12.53 & 12.75 & 12.01 & 12.64 &  14.43 &  14.50 \\
Mela(ours)        & 800M  & 10.50 & 10.75 & 10.04 & 10.82 &  12.74 &  13.41 \\
Mela(ours)        & 1.2B  &  9.08 &  9.49 &  9.20 &  9.91 &  11.64 &  11.85 \\
\bottomrule
\end{tabular}
\caption{Perplexity (PPL) across context lengths for Mela and Transformer models.}
\label{tab:ppl_results}
\end{table}

\section{Ablation Studies}

\end{document}